\documentclass[letterpaper, 10 pt, conference]{ieeeconf}  

\IEEEoverridecommandlockouts                              

\usepackage{graphics} 
\usepackage{booktabs}
\usepackage{graphicx}
\usepackage{multirow}
\usepackage{float}
\usepackage{mathrsfs}
\usepackage{colortbl}
\usepackage{color}
\usepackage[table]{xcolor} 

\definecolor{mygray}{gray}{.88}
\usepackage{bbding}
\usepackage{caption}
\usepackage{pdfpages}
\usepackage{makecell}
\usepackage{amssymb}

\usepackage[backend=biber,style=numeric,sorting=none]{biblatex}
\usepackage{marvosym}
\usepackage{etoolbox}

\usepackage{amsmath}
\usepackage{algorithm}
\usepackage[noend]{algpseudocode}
\usepackage{graphicx}
\usepackage{stfloats}
\usepackage{subfigure}
\usepackage{caption}
\usepackage{setspace}
 
\usepackage{lipsum}
\usepackage{caption}
\usepackage{indentfirst} 
\usepackage{hyperref}
\hypersetup{hypertex=true,
colorlinks=true,
linkcolor=blue,
anchorcolor=blue,
citecolor=black}

\usepackage{amsmath} 
\definecolor{lightblue}{rgb}{0.8,0.85,1} 

\definecolor{darkblue}{rgb}{0.414902, 0.561765, 0.982353}

\newcommand\blfootnote[1]{%
\begingroup 
\renewcommand\thefootnote{}\footnote{#1}%
\addtocounter{footnote}{-1}%
\endgroup }
\AtEveryBibitem{\small}
\DeclareFieldFormat{labelnumberwidth}{\footnotesize\mkbibbrackets{#1}}

\defbibenvironment{bibliography}
  {\list
     {\printfield[labelnumberwidth]{labelnumber}}
     {
      \setlength{\labelwidth}{\labelnumberwidth}%
      \setlength{\leftmargin}{\labelwidth}%
      \setlength{\labelsep}{\biblabelsep}%
      \addtolength{\leftmargin}{\labelsep}%
      \setlength{\itemsep}{\bibitemsep}%
      \setlength{\parsep}{\bibparsep}}}
  {\endlist}
  {\item}
\begin{document}
\title{\LARGE \bf
Multi-Floor Zero-Shot Object Navigation Policy 
}

\author{Lingfeng Zhang$^{1*}$, Hao Wang$^{1*}$,  Erjia Xiao$^{1*}$, Xinyao Zhang$^{1}$, Qiang Zhang$^{1,2}$, Zixuan Jiang$^{1}$, Renjing Xu$^{1,\dag}$
\thanks{$^{1}$Authors with The Hong Kong University of Science and Technology\newline(Guangzhou). {\tt\small lzhang819@conncet.hkust-gz.edu.cn}}
\thanks{$^{2}$Author with Beijing Innovation Center of Humanoid Robotics Co., Ltd.
{\tt\small  Jony.Zhang@x-humanoid.com}}%
\thanks{$^\dag$ is the Corresponding Author. {\tt\small renjingxu@hkust-gz.edu.cn}}
}





\thispagestyle{empty}
\pagestyle{empty}
\twocolumn[{
\renewcommand\twocolumn[1][]{#1}
\maketitle
\vspace{-20pt}
\begin{center}
    \captionsetup{type=figure}
    \includegraphics[width=1.0\textwidth]{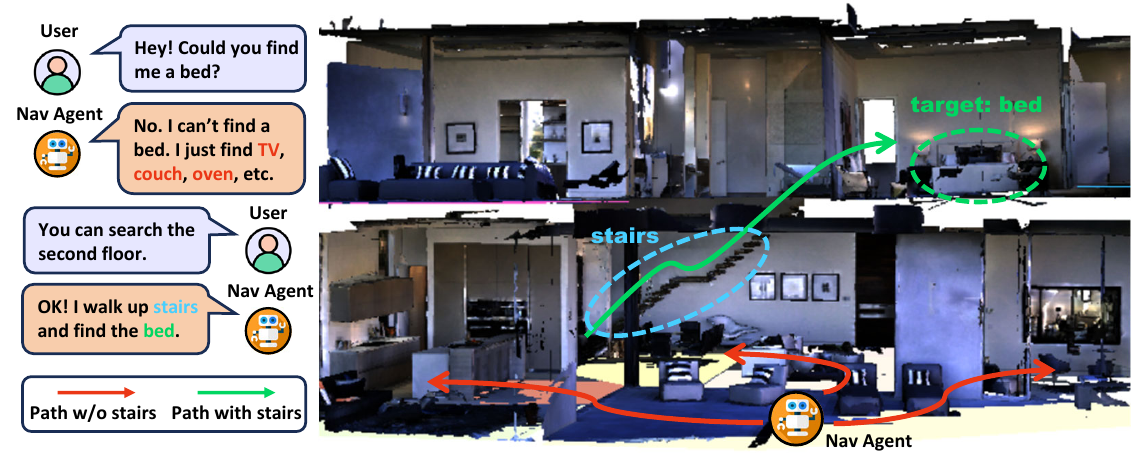}
    \vspace{-10pt}
    \captionof{figure}{\textbf{The challenge of single-floor navigation.} The target object of indoor object navigation is very likely to appear on different floors of the house, so the agent may not find the target object even if it has fully explored the current floor. Our proposed stair policy introduces the concept of indoor multi-floor navigation and proposes a feasible and learning-free solution to this challenge.}
    \label{challenge}
\end{center}

}]

\blfootnote{$^{1}$Authors with The Hong Kong University of Science and Technology (Guangzhou). $^{\ast}$ Joint first authors.    
         {\tt\small lfzhang715@gmail.com}\\
\indent $^{2}$Author with Beijing Innovation Center of Humanoid Robotics Co., Ltd.\\
{\tt\small  Jony.Zhang@x-humanoid.com}\\
\indent $^\dag$ is the Corresponding Author. {\tt\small renjingxu@hkust-gz.edu.cn}
}

\vspace{-0.5cm}
\begin{abstract}

Object navigation in multi-floor environments presents a formidable challenge in robotics, requiring sophisticated spatial reasoning and adaptive exploration strategies. Traditional approaches have primarily focused on single-floor scenarios, overlooking the complexities introduced by multi-floor structures. To address these challenges, we first propose a Multi-floor Navigation Policy (MFNP) and implement it in Zero-Shot object navigation tasks. Our framework comprises three key components: (i) Multi-floor Navigation Policy, which enables an agent to explore across multiple floors; (ii) Multi-modal Large Language Models (MLLMs) for reasoning in the navigation process; and (iii) Inter-Floor Navigation, ensuring efficient floor transitions. We evaluate MFNP on the Habitat-Matterport 3D (HM3D) and Matterport 3D (MP3D) datasets, both include multi-floor scenes. Our experiment results demonstrate that MFNP significantly outperforms all the existing methods in Zero-Shot object navigation, achieving higher success rates and improved exploration efficiency. Ablation studies further highlight the effectiveness of each component in addressing the unique challenges of multi-floor navigation. Meanwhile, we conducted real-world experiments to evaluate the feasibility of our policy. Upon deployment of MFNP, the Unitree quadruped robot demonstrated successful multi-floor navigation and found the target object in a completely unseen environment. By introducing MFNP, we offer a new paradigm for tackling complex, multi-floor environments in object navigation tasks, opening avenues for future research in vision-based navigation in realistic, multi-floor settings.

\end{abstract}

\section{INTRODUCTION}
\vspace{-0.2cm}
Navigating in unknown environments to find specified target objects remains a significant challenge in Embodied AI research. The Habitat Challenge, through its indoor Object Goal Navigation (ObjectNav) task benchmark, aims to assess the ability of agents to locate specific objects (e.g., bed, TV monitor) within complex 3D indoor scenes\cite{batra2020objectnav}. In this task, agents must navigate using only information captured by an RGB-D camera and global pose data.\\
\indent Recent years have witnessed the development of various navigation methodologies for ObjectNav, including reinforcement learning\cite{mazoure2022improving}\cite{filos2021psiphi}\cite{deitke2022️}, imitation learning\cite{cai2023bridging}\cite{chaplot2020object}, Zero-Shot learning\cite{zhang2021hierarchical}\cite{luo2022stubborn}\cite{zhou2023esc}, and Few-Shot learning\cite{yu2023l3mvn}.  Reinforcement learning has shown promise in training agents to make sequential decisions, allowing them to learn optimal navigation policies through trial and error in simulated environments. Imitation learning significantly improves task success by teaching agents to navigate as humans do but requires extensive training and human demonstrations. Zero-shot and Few-Shot methods offer advantages in deployability and adaptability to different scenes, despite a slight decrease in accuracy compared to fully trained models.\\
\indent However, these approaches have primarily focused on single-floor scenarios, overlooking the complexities of multi-floor navigation in indoor environments. This omission represents a critical gap in current ObjectNav research. In real-world scenarios, target objects are often distributed across multiple floors of a building. For instance, bedroom furniture like beds are more likely to be found on upper floors, while living room items such as sofas are typically located on ground floors. This spatial distribution of objects across different levels introduces additional challenges that existing navigation methods have not adequately addressed.

\indent To bridge this gap, we first propose a novel approach that incorporates the Multi-floor Navigation Policy (MFNP) into Zero-Shot object navigation methods. Our work is motivated by the observation that target objects are distributed on different floors in multi-floor indoor environments.\\

\vspace{-7pt}
\indent The main contributions of our study are as follows:

\begin{itemize}
\item We conduct a comprehensive analysis of ObjectNav scenarios where target objects are located on different floors, quantifying the frequency and impact of these multi-floor navigation challenges.
\item We first propose a novel multi-floor navigation policy for ObjectNav, specifically designed to enable agents to efficiently navigate between floors through stairs.
\item We integrate our multi-floor navigation policy into a Zero-Shot learning framework, significantly enhancing the success rate of ObjectNav tasks in complex, multi-floor environments.
\end{itemize} 

\indent Through extensive experimentation, we demonstrate that our approach achieves state-of-the-art (SOTA) performance among all the Zero-Shot methods for ObjectNav on the Habitat platform. Importantly, we have also conducted real-world experiments, validating the effectiveness of our policy in real-world environments. 
These results underscore the importance of considering multi-floor navigation in the design of vision-based navigation and pave the way for future research in this critical area of indoor navigation.





\section{RELATED WORK}

\subsection{Object Navigation}

Visual navigation is a critical task for robots, especially in unknown environments. Object Goal Navigation (ObjectNav) focuses on visual navigation within these unknown settings, leveraging semantic priors to enhance a robot's ability to locate objects\cite{batra2020objectnav}. Implementations of the ObjectNav often rely on reinforcement learning\cite{mazoure2022improving,filos2021psiphi,deitke2022️}, imitation learning\cite{cai2023bridging}, or top-down map predictions\cite{chaplot2020object,zhang2021hierarchical,luo2022stubborn}. However, these methods are predominantly based on closed dataset research, making them less applicable to different datasets and platforms. To address the challenges of applying the ObjectNav to various datasets and reduce training consumption, recent developments in Zero-Shot ObjectNav frameworks have garnered significant attention. We will also employ Zero-Shot approaches to conduct MFNP.

\subsection{Large Models for Object Navigation}

The emergence of large models, including Large Language Models (LLMs) and Multi-modal Large Models (MLLMs), trained on Internet-scale datasets, has introduced powerful abilities such as planning, reasoning and analyzing. These abilities are particularly relevant to object navigation tasks that require the use of a variety of high-level information, motivating their use in ObjectNav.
Recent studies have explored various ways to exploit these models:\cite{zhou2023esc}\cite{yokoyama2023vlfm}\cite{huang2023visual}
directly utilizing the visual perceptual capabilities of multi-modal large-scale models to aid the exploration process. Other approaches such as \cite{yu2023l3mvn}\cite{zhang2024trihelper} utilize large language models for high-level navigation policy.

While these approaches take advantage of the powerful generalization capabilities of large models, none of them use large models for multi-floor navigation planning. Our proposed framework MFNP introduces a new approach to multi-floor navigation policy planning using LLMs. This approach enables the system to process and interpret various data in navigation to generate hierarchical navigation plans, aiming to solve the indoor multi-floor navigation problem.

\begin{figure*}[!htbp]
\centering
\includegraphics[width=1.0\linewidth]{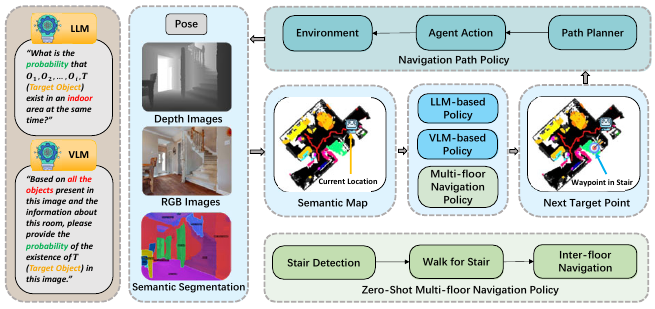}
\caption{\textbf{The general pipeline of our framework.} Firstly, we construct a semantic map using RGB-D observations $V_t$ and global pose $G_t$. Then we obtain various information from the semantic map and input it into policies to obtain the next waypoint. Our proposed stair policy will make the exploration decision to other floors and guide the agent throughout the process. After obtaining the next waypoint, we use the path planning policy to calculate the final action.}
\vspace{-0.5cm}
\label{framework}
\end{figure*}

\label{sec:frontier}
\section{PRELIMINARY}

\subsection{Problem Formulation} 

The ObjectNav task requires agents to locate and approach predefined target object categories in unseen environments. The task defines a set of target categories $T$, such as "chair" or "bed". At the beginning of each episode, the agent is randomly initialized at a location within scene $S$ and is assigned a specific target category $T_i$.

At each discrete time step $t$, the agent receives an observation vector $O_t = (V_t, P_t)$, where $V_t$ represents visual input (including RGB images and depth maps), and $P_t$ represents the agent's pose information. Based on these inputs, the agent must select an action $a_t \in \mathcal{A}$, where the action space $\mathcal{A}$ contains six discrete actions: move forward, turn left, turn right, look up, look down, and stop. When the agent determines it has approached the target object, it can choose to execute the stop action.


Task success is defined as follows: when the agent actively stops, its Euclidean distance from the target object must be less than a predetermined threshold $d$ (set to 0.1 meters in this task). Each navigation episode is limited to 500 time steps.

\subsection{Semantic Map}
The construction of the semantic map $\mathcal{M}$ employs the method proposed by \cite{chaplot2020object}, utilizing RGB-D images $V_t$ and the agent's ground truth pose data $P_t$. The map is populated by converting visual data into point clouds using geometric algorithms, which are then projected onto a 2D top-down view. This approach incorporates physical obstacles, explored areas, and semantically segmented object categories. Precise alignment between semantic masks and point clouds enables accurate channel mapping within the semantic map. The map is represented as a three-dimensional tensor with dimensions $C \times W \times H$, where $W$ and $H$ denote the map's width and length, respectively, and $C$ equals $C_n + 5$, with $n$ representing the number of object categories. The tensor's initial four channels encode obstacle information, explored terrain, current agent position, and historical agent locations. The subsequent $n$ channels delineate semantic maps for $n$ distinct object types, followed by an additional channel dedicated to stair mapping for subsequent analysis of stair navigation policy. At the beginning of each episode, the semantic map is initialized, with the agent's starting position defaulting to the map's central coordinates. Semantic maps serve as the foundational element that enables our system effectively navigate to target objects without requiring any prior training on the specific instances.

\subsection{Candidate Waypoints Map}
After constructing the semantic map, we generate a set of candidate exploration points derived from the first two channels of the semantic map, following a methodology similar to that outlined in \cite{ramakrishnan2022poni}. The process begins by identifying the boundaries of the explored area to determine the outer perimeter. We then expand the edges of the obstacle map and subtract it from the explored area to highlight potential exploration targets. Small, insignificant areas are filtered out, leaving only substantial regions as viable candidates for exploration. The centroids of these remaining areas constitute our set of candidate points, denoted as $\mathcal{P}$.

To prioritize these candidate points, we employ a scoring system based on cost and utility, adapted from \cite{ramakrishnan2022poni}. For each candidate point $ p_i \in \mathcal{P}$, we calculate a score $S(p_i)$ using the following equation:
\begin{equation}
S(p_i) = B(p_i) - \alpha D(p_i)
\label{eq:score}
\end{equation}

where $B(p_i)$ represents the benefit function, $D(p_i)$ is the distance function (serving as a proxy for cost), and $\alpha$ is a constant that adjusts the relative importance between these two factors. Each potential candidate point is evaluated to determine its viability as an exploration destination, balancing the cost of reaching the point with the expected benefit of exploration.
\begin{figure*}[!htbp]
\centering
\includegraphics[width=1.0\linewidth]{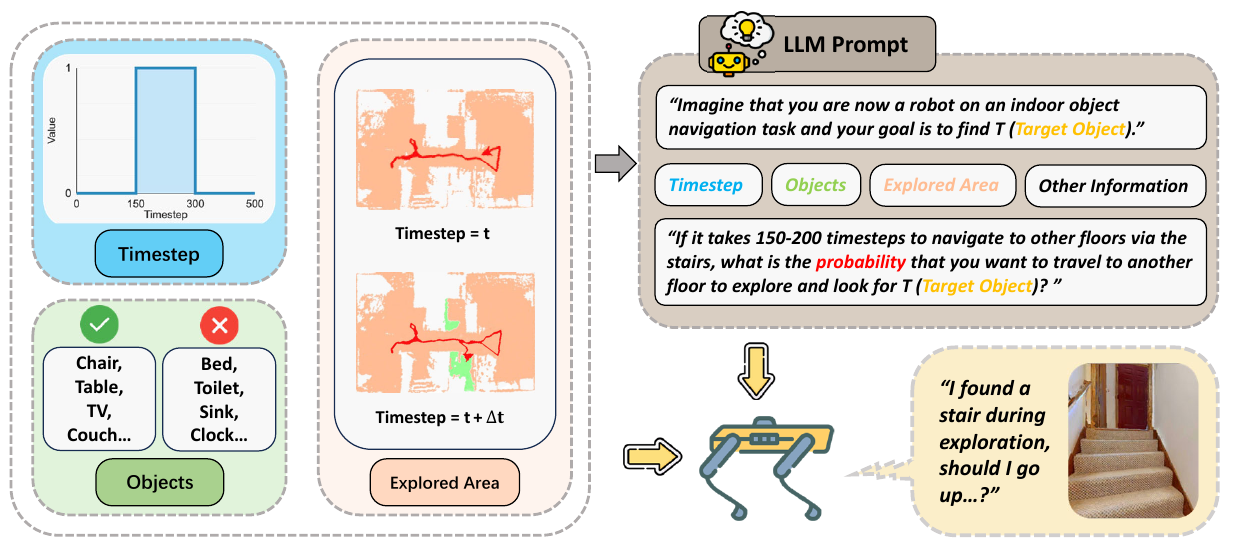}
\caption{The architecture of our Multi-floor Navigation Policy. We aggregate and maintain a prompt encompassing the exploration information from each timestep to elicit recommendations from the LLM. Subsequently, we synthesize and weight all acquired information to arrive at a final determination.}
\vspace{-0.5cm}
\label{modules}
\end{figure*}
\label{sec:MFNP}

\section{METHODOLOGY}
\vspace{-0.15cm}
\subsection{Pipeline}
The general pipeline of our proposed framework is illustrated in Fig. \ref{framework}. 
After obtaining the RGB-D image $V_t$ and agent pose $P_t$ from the simulator's environment observation, we input $V_t$ into the semantic segmentation module to obtain $S_t$, and thus use $V_t, P_t,$ and $S_t$ to construct the semantic map. After constructing the semantic map, we use a series of policies including LLM-based, VLM-based, and MFNP to select the next waypoint. Among them, LLM-based Policy is mainly responsible for candidate point selection, while VLM-based Policy for target object detection, and MFNP for multi-floor exploration. After that, we use the navigation path policy mentioned in \ref{sec:policy} to calculate the agent's action for point-to-point navigation to the waypoint. At each timestep, we calculate a new action for the agent and interact with the environment to collect data for the next timestep. Each policy will be described in detail below, and dynamically using these policies during navigation ensures that our agent can efficiently perform multi-floor ObjectNav.

\subsection{Multi-floor Navigation Policy}

\subsubsection{LLM-based Policy}
After obtaining the set of candidate points $\mathcal{P}$, we use an LLM-based policy for candidate point selection. As shown in the LLM prompt on the left side of Fig. \ref{framework}, we obtain scores for each candidate point $p_i$ and sort them to obtain the set of scores $\mathcal{G}$ for the candidate points. To prevent the local optimal problem of LLM selection and improve the exploration efficiency, we calculate the size of the proportion of explored area around each candidate point $R_i = \frac{Area_{explored}}{Area_{unexplored}} $. if $R_i$ is greater than 90\%, we exclude this candidate point and select the second highest rated candidate point in $\mathcal{G}$ as the next waypoint.  At the same time, we introduce the method proposed in \cite{zhang2024trihelper} to detect the case that the LLM selects the same candidate point repeatedly, which will allow the agent to explore freely to reduce the unexplored area.

\subsubsection{VLM-based Policy}
Target object detection is a critical component in ObjectNav. To optimize this process and mitigate potential misclassifications by semantic segmentation modules (e.g., misidentifying a mural as a television), we propose a VLM-based policy, inspired by the approach in \cite{zhang2024trihelper}.
The core concept of our method is to employ a Vision-Language Model (VLM) to perform a double-check on the current frame, assessing the likelihood of the target object's presence. The prompt used for the VLM is illustrated in Fig. \ref{framework}.
Upon obtaining the probability from the VLM, we integrate it with the confidence score from the semantic segmentation module through a weighted combination. This integration can be expressed as:
\begin{equation}
C_{conf} = \beta * P_{seg} + (1 - \beta) * P_{vlm}
\label{eq:weight}
\end{equation}
where $C_{conf}$ is the final confidence score for decision-making, $P_{seg}$ is the confidence score from semantic segmentation, $P_{vlm}$ is the probability obtained from the VLM, $\alpha$ is a weighting factor ($0 \leq \beta \leq 1$). 
We leverage the contextual understanding and multimodal capabilities of VLM to enhance the robustness of object recognition in navigation tasks. By combining the strengths of semantic segmentation and VLM, we aim to reduce false positives in target detection and improve overall navigation accuracy.

\subsubsection{Multi-floor Navigation Policy}

In this section, we elucidate our policy for implementing multi-floor object navigation in indoor environments: MFNP. This approach leverages various information sources and the cognitive capabilities of Large Language Models (LLMs).

A fundamental component of our policy is the determination of staircase existence, denoted as $E_{stair}$. We employ the semantic segmentation module to identify staircases and map their locations onto a semantic map for localization. $E_{stair}$ is binary, with $1$ indicating the presence of a staircase and $0$ signifying its absence. The MFNP is activated only when the staircase is present, i.e. $E_{stair}=1$.

To establish the optimal timing for multi-floor exploration, we aggregate information collected by the agent during its current exploration phase. This information encompasses three key metrics: timestep, objects, and explored area.

Regarding the timestep, each episode is constrained to a maximum of $500$ timesteps. To ensure efficacy and allocate sufficient time for inter-floor navigation, we prohibit the agent from executing stair ascent during the initial $150$ timesteps and the final $200$ timesteps. Additionally, we implement a time-dependent validity function $f(t)$ for the timestep, which decreases as the episode progresses.

For the objects metric, we enumerate all explored object categories and calculate the ratio of discovered objects ($O_{explored}$) to the total number of object types ($O_{total}$). The greater the variety of objects that have been explored, the lower the probability that the target object will be present in that floor.

To assess the explored area, we monitor the change in exploration coverage ($E_t$) over fixed time intervals ($\Delta t $). This is accomplished by recording the explored area at the beginning of each interval and computing the proportional change. A smaller proportion indicates a higher degree of exploration on the current floor, suggesting that exploration of other floors may be more beneficial.
\begin{figure*}[!htbp]
\centering
\includegraphics[width=1.0\linewidth]{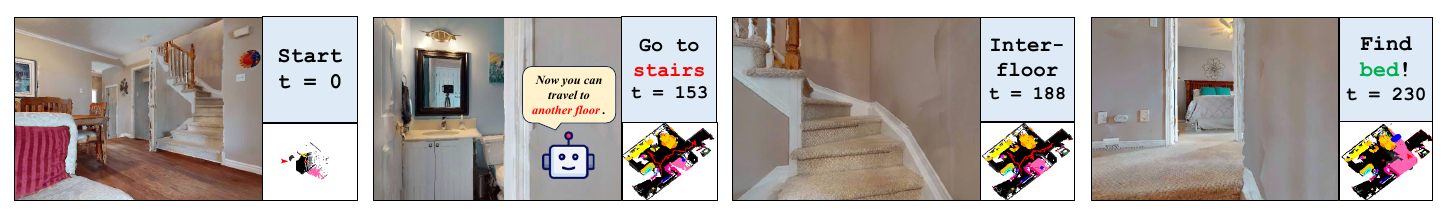}
\caption{Our policy proceeds on the Habitat platform.}
\vspace{-0.7cm}
\label{modules}
\end{figure*}
\label{sec:example1}

Upon obtaining these three critical metrics, we maintain a prompt as illustrated in Fig.  \ref{modules}. The relevant information is encapsulated in the following prompt structure:

\textit{"Your task has a time limit of 500 timesteps, and the current timestep is $t$. On the present floor, you have identified objects $O_1, O_2, ..., O_n$ out of a total of $O_{total}$ object types. In the past $\Delta t $ timesteps, the proportion of newly explored area is $E_t$."}

So when $E_{stair}=1$ and timestep is within a certain range, we can then synthesize to get the judgment metrics for the multi-floor navigation policy:
\begin{equation}
N_{\text{MFNP}} = w_1 f(t) + w_2 \frac{O_{\text{explored}}}{O_{\text{total}}} + w_3(1-E_t) + w_4 P_{\text{LLM}}
\label{eq:strategy}
\end{equation}

where $f(t)=\frac{500-t}{500}$ denotes the time-dependent validity function for timestep. $\frac{O_{\text{explored}}}{O_{\text{total}}}, E_t, P_{LLM}$ represent the respective information, we weigh each parameter and get the final $N_{\text{MFNP}}$. If $N_{\text{MFNP}}$ is larger than a certain threshold $N_{th}$ we will tell the agent to explore another floor via the stairs.

\subsubsection{Inter-floor Navigation}
Upon the agent's decision to move to a different floor, we implement a comprehensive inter-floor navigation policy to facilitate efficient stair traversal. This policy begins by designating the stair area as the next waypoint, guiding the agent toward this critical transition point. Once the agent accesses the stair area, we employ a dilation operation on the stair entrance area and subsequently mark it as an obstacle, a crucial step that prevents the agent from retracing its path and effectively enforces unidirectional movement through the staircase. By constraining the agent's options to a single exit path on the opposite end of the stair, we can define a specific waypoint, allowing the agent to utilize the path planner mentioned in \ref{sec:policy} for effective inter-floor navigation. To address scenarios where the agent either fails to reach the next floor or has already transitioned to it without updating its map, we implement a time-based reset mechanism. After a predetermined duration of 200 timesteps, the semantic map is reinitialized, and the multi-floor navigation policy is temporarily disabled, allowing the agent to resume exploration. This comprehensive approach ensures robust and efficient inter-floor navigation.


%


\subsection{Navigation Path Policy}

After finding the next waypoint through the previous policies, we utilize a path-planning policy to navigate the agent to the goal point. We use the method in \cite{sethian1996fast} for point-to-point navigation from the current position to the next waypoint. Path planner calculates the point navigation path at each timestep and gives agent the action to be performed from the computed path. This point-to-point path planner makes our pipeline independent of end-to-end training, while providing a high-level solution for inter-floor navigation.
\label{sec:policy}

\section{EXPERIMENT}

\vspace{-0.2cm}
\subsection{Datasets}
We use the simulator \cite{savva2019habitat} to evaluate our approach on two datasets, HM3D\cite{ramakrishnan2021habitat} and MP3D\cite{chang2017matterport3d}, both of which are multi-floor scenes that can be well laid out for our multi-floor navigation policy. Specifically, the validation segment for HM3D includes 2000 episodes and \textbf{6} target object categories, spread over 20 scenes. MP3D’s validation split contains 2195 episodes across 11 scenes and \textbf{21} target object categories. 

\indent We quantified the proportion of scenes in MP3D where target object categories are distributed across multiple floors, necessitating multi-floor exploration for ObjectNav. It indicates that \textbf{57.3\%} of the scenes in the MP3D dataset contain object categories that are not confined to a single floor, which demonstrates the importance of our policy. It is worth noting that while our analysis focused on the MP3D, we were unable to conduct a similar assessment for HM3D due to the absence of floor annotation information.

\vspace{-5pt}
\subsection{Experiment Details}
\vspace{-2pt}
We evaluated our policy on the Habitat platform 2.0\cite{savva2019habitat}. Our implementation is based on the architecture of \cite{yu2023l3mvn} and \cite{zhang2024trihelper}. The LLM and VLM used for the experiments are Qwen2\cite{yang2024qwen2} and Qwen2-VL-Int4\cite{bai2023qwen}, respectively. Meanwhile, we use Mask2Former\cite{cheng2022masked} for semantic segmentation to predict all existing objects in the RGB-D image.
\vspace{-5pt}
\subsection{Metrics}
\vspace{-5pt}
Our approach is assessed utilizing metrics established in prior research by \cite{anderson2018evaluation}. These metrics include Success Rate (SR), Success weighted by Path Length (SPL), and Distance to Goal (DTG).  In this evaluation framework, higher values for SR and SPL indicate superior performance. The SPL incorporates both task completion and path efficiency, comparing the agent's actual trajectory length to the optimal path length. Conversely, DTG quantifies the terminal distance between the agent and target objects at episode conclusion, with lower values signifying better outcomes.
\vspace{-5pt}
\subsection{Baselines}
\vspace{-2pt}
To evaluate the Zero-Shot ObjectNav performance of our model, we compare it to several baselines containing the state-of-the-art (SOTA) baseline. L3MVN\cite{yu2023l3mvn} used LLM to select the candidate waypoints. VLFM\cite{yokoyama2023vlfm} built a semantic value map to evaluate frontiers to select exploration directions. And TriHelper\cite{zhang2024trihelper} proposed three helpers to solve three main challenges in Zero-Shot ObjectNav.

\subsection{Results}
The performance of our model on the two datasets HM3D and MP3D is shown in Table \ref{comp}. MFNP outperforms all the Zero-Shot ObjectNav methods and achieves SOTA on both datasets. Compared to previous SOTA work, our proposed method achieves +1.8\% SR improvement and +5.5\% SPL on the HM3D dataset; +4.7\% SR improvement on the MP3D dataset. 
The reason why our SPL is slightly lower than that of the SOTA method is that the proposed Multi-floor policy focuses on exploring extra floors, which causes the agent search path to be much longer in some episodes.
\renewcommand{\arraystretch}{1.0}
\begin{table}[ht]
\vspace{-0.2cm}
\scriptsize
\caption{Results of comparative experiment.}
    \centering
    \setlength{\tabcolsep}{1.4mm}{
    \begin{tabular}{cccccccc}
    \toprule[1.0pt]
    
        \multirow{2}{*}{\textbf{Method}} & \multirow{2}{*}{\textbf{Zero-Shot}} & \multicolumn{3}{c}{\textbf{HM3D}}  & \multicolumn{3}{c}{\textbf{MP3D}} \\ \cmidrule(r){3-5} \cmidrule(l){6-8} 
        ~ & ~ & SR$\uparrow$ & SPL$\uparrow$ &DTG$\downarrow$ & SR$\uparrow$ & SPL$\uparrow$ & DTG$\downarrow$ \\ \midrule[1.0pt]
        ZSON\cite{majumdar2022zson} & \XSolidBrush & 25.5 & 0.126 & - & 15.3 & 0.048 & - \\ 
        PONI\cite{ramakrishnan2022poni} & \XSolidBrush & - & - & - & 31.8 & 0.121 & 5.1 \\
        PixNav\cite{cai2023bridging} & \XSolidBrush & 37.9 & 0.205 & - & - & - & - \\ 
        SPNet\cite{zhao2023semantic} & \XSolidBrush & 31.2 & 0.101 & - & 16.3 & 0.048 & - \\ \midrule[1.0pt]
        
        CoW\cite{gadre2023cows} & \Checkmark  & - & - & - & 7.4 & 0.037 & -\\ 
        ESC\cite{zhou2023esc} & \Checkmark  & 39.2 & 0.223 & - & 28.7 & 0.142 & - \\

        VLFM\cite{yokoyama2023vlfm} & \Checkmark & 52.5 & \textbf{0.304} & - & 36.4 & \textbf{0.175} & - \\ 
        
        VoroNav\cite{wu2024voronav} & \Checkmark  & 42.0 & 0.260 & - & - & - & - \\
        L3MVN\cite{yu2023l3mvn}& \Checkmark  & 50.4 & 0.231 & 4.427 & - & - & - \\
        TriHelper\cite{zhang2024trihelper}& \Checkmark  & 56.5 & 0.253 & 3.873 & - & - & - \\ 
        InstuctNav\cite{long2024instructnav}& \Checkmark  & 56.0 & 0.225 & - & - & - & - \\ \midrule[1.0pt]
        \rowcolor{mygray}\textbf{MFNP (Ours)} & \Checkmark & \textbf{58.3} & 0.267&\textbf{3.568} &\textbf{41.1} & 0.154 & \textbf{4.53} \\ 

    \bottomrule[1.0pt]
    \end{tabular}}
    \label{comp}
\vspace{-0.35cm}
\end{table}

\subsection{Ablation study}

    
        

\begin{figure*}[!ht]
\centering
\includegraphics[width=1\linewidth]{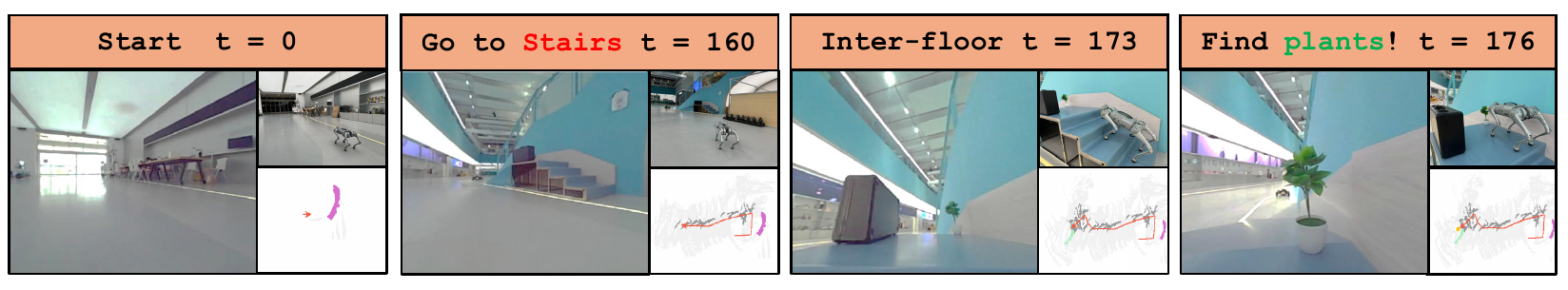}
\caption{Real-world demonstration (a Unitree quadruped robot) of MFNP}
\vspace{-0.5cm}
\label{appendix1}
\end{figure*}

We conducted ablation experiments on each informational component of the MFNP algorithm across all episodes in the HM3D dataset, demonstrating the efficacy of each component, the results are shown in Table \ref{ablation}. It can be seen that each component is integral to the use of MFNP and works best when used in concert. Additionally, we randomly selected a subset of 400 episodes where MFNP was triggered, indicating the agent's decision to engage in multi-floor exploration. This curated validation set was then used to evaluate baseline methods. From the results, we can see that our policy of having the agent actively explore other floors in this subset is effective.
\subsection{Real-world Implement}
We set up our real-world experiments on the Unitree quadruped robot Go1-Edu. We used the binocular fisheye camera on the robot and IMU to acquire RGB and depth images as well as pose information. To speed up processing, we uploaded the data to the cloud and used a RTX 4090 workstation to compute and return motions and the Go1 SDK for droid control. Here we use the climbing stair SDK for the up-stair action to solve the inter-floor navigation problem. To ensure realism, we used the same model configuration as well as parameters for inference and only changed the input image size. Fig. \ref{appendix1} shows one of our successes, Go1 goes up the stairs and find the target object.
\renewcommand{\arraystretch}{1.0}
\begin{table}[h]
\scriptsize
\caption{Results of ablation study on MP3D.}
    \centering
    \resizebox{\linewidth}{!}{
    \begin{tabular}{lccccc}
    \toprule[1.0pt]
    
        \textbf{Methods}& \textbf{Episodes} & \textbf{SR}$\uparrow$   & \textbf{SPL}$\uparrow$ & \textbf{DTG}$\downarrow$  &\textbf{Multi-floor}\\ \midrule
    
        \textbf{MFNP} &        ALL    &  \textbf{58.3} & \textbf{0.267} & \textbf{3.568}    & \Checkmark \\ \midrule
        w/o Timestep&    ALL         &  55.8 & 0.231 & 3.891   &\Checkmark \\
        w/o Objects&    ALL     &   56.2  & 0.251 & 3.714  & \Checkmark \\
        w/o Explored&   ALL      & 54.0  & 0.210  & 4.032 &\Checkmark \\ 
        w/o LLM&       ALL   &  57.1  &   0.260 & 3.661&\Checkmark  \\  \midrule
         L3MVN\cite{yu2023l3mvn} &     Subset     &   27.6  & 0.113& 6.032  & \XSolidBrush  \\
         TriHelper\cite{zhang2024trihelper}&  Subset        &   33.0 & 0.132& 5.687   & \XSolidBrush\\
         \textbf{MFNP} &        Subset    &  \textbf{38.5} & \textbf{0.141} & \textbf{5.359}    & \Checkmark \\ 
    \bottomrule[1.0pt]
    \end{tabular}}
\label{ablation}
\end{table}

\section{CONCLUSIONS}


In this work, we presented Multi-floor Navigation Policy (MFNP), an innovative framework designed to address the multi-floor challenges in ObjectNav. Our framework significantly enhances the navigational capabilities of autonomous agents operating within unseen multi-floor settings by dynamically integrating three key components: Multi-floor Navigation Policy, Multi-modal Large Language Models (MLLMs) Reasoning, and Inter-floor Navigation. Our comprehensive experiments conducted on the HM3D and MP3D datasets have demonstrated the superior performance of MFNP compared to all existing Zero-Shot ObjectNav methods. The ablation studies further validated the critical role of each component, particularly highlighting the effectiveness of multi-floor exploration. \\
\indent We first notice the challenge of multi-floor navigation and propose a corresponding policy MFNP to address it. The importance of considering vertical spatial relationships based on horizontal exploration is demonstrated. Our work opens up new perspectives for future research on vision-based navigation in realistic multi-floor environments.\\
\indent In the future, to further optimize navigation system, we can make deeper extensions to the multi-floor navigation policy, such as expanding to refine the sparse multi-floor navigation dataset and utilizing it for end-to-end training-based methods like reinforcement learning and imitation learning.





\printbibliography

\clearpage

\end{document}